\documentclass[letterpaper]{article} % DO NOT CHANGE THIS
\usepackage{aaai24}  % DO NOT CHANGE THIS
\usepackage{times}  % DO NOT CHANGE THIS
\usepackage{helvet}  % DO NOT CHANGE THIS
\usepackage{courier}  % DO NOT CHANGE THIS
\usepackage[hyphens]{url}  % DO NOT CHANGE THIS
\usepackage{graphicx} % DO NOT CHANGE THIS
\usepackage{natbib}  % DO NOT CHANGE THIS AND DO NOT ADD ANY OPTIONS TO IT
\usepackage{caption} % DO NOT CHANGE THIS AND DO NOT ADD ANY OPTIONS TO IT
\usepackage{algorithm}
\usepackage{algorithmic}

\usepackage{multirow}
\usepackage[table,xcdraw]{xcolor}
\usepackage{makecell}
\usepackage{environ}         % provides \BODY
\usepackage{etoolbox}        % provides \ifdimcomp
\usepackage{soul}
\usepackage{subcaption}
\usepackage{mathrsfs}        % provides \mathscr
\usepackage{amsmath}
\usepackage{amssymb}
\newlength{\myl}
\let\origequation=\equation
\let\origendequation=\endequation

\RenewEnviron{equation}{
  \settowidth{\myl}{$\BODY$}                       % calculate width and save as
  \myl
  \origequation
  \ifdimcomp{\the\linewidth}{>}{\the\myl}
  {\ensuremath{\BODY}}                             % True
  {\resizebox{\linewidth}{!}{\ensuremath{\BODY}}}  % False
  \origendequation
}

\usepackage[capitalize]{cleveref}
\crefname{section}{Sec.}{Secs.}
\Crefname{section}{Section}{Sections}
\Crefname{table}{Table}{Tables}
\crefname{table}{Tab.}{Tabs.}

\usepackage{newfloat}
\usepackage{listings}
\DeclareCaptionStyle{ruled}{labelfont=normalfont,labelsep=colon,strut=off} % DO NOT CHANGE THIS
\floatstyle{ruled}
\newfloat{listing}{tb}{lst}{}
\floatname{listing}{Listing}
\title{CPN: Complementary Proposal Network for Unconstrained Text Detection}
\author{
    Longhuang Wu,
    Shangxuan Tian\thanks{Corresponding author},
    Youxin Wang,
    Pengfei Xiong
}
\affiliations{
    Shopee Pte. Ltd.\\
    \{wlonghuang, wyxileroy, xiongpengfei2019\}@gmail.com,
    tianshangxuan@u.nus.edu
}

\usepackage{bibentry}
\begin{document}

\maketitle

\begin{abstract}
Existing methods for scene text detection can be divided into two paradigms: segmentation-based and anchor-based. While Segmentation-based methods are well-suited for irregular shapes, they struggle with compact or overlapping layouts. Conversely, anchor-based approaches excel for complex layouts but suffer from irregular shapes. To strengthen their merits and overcome their respective demerits, we propose a Complementary Proposal Network (CPN) that seamlessly and parallelly integrates semantic and geometric information for superior performance. The CPN comprises two efficient networks for proposal generation: the Deformable Morphology Semantic Network, which generates semantic proposals employing an innovative deformable morphological operator, and the Balanced Region Proposal Network, which produces geometric proposals with pre-defined anchors. To further enhance the complementarity, we introduce an Interleaved Feature Attention module that enables semantic and geometric features to interact deeply before proposal generation. By leveraging both complementary proposals and features, CPN outperforms state-of-the-art approaches with significant margins under comparable computation cost. Specifically, our approach achieves improvements of 3.6\%, 1.3\% and 1.0\% on challenging benchmarks ICDAR19-ArT, IC15, and MSRA-TD500, respectively. Code for our method will be released.
\end{abstract}

\section{Introduction}
\label{sec:intro}

\begin{figure}[t]
 % \centering
 \centering
 \setlength{\abovecaptionskip}{0.cm}
   \begin{subfigure}{0.49\linewidth}
       \includegraphics[width=\linewidth, height=0.6\linewidth]{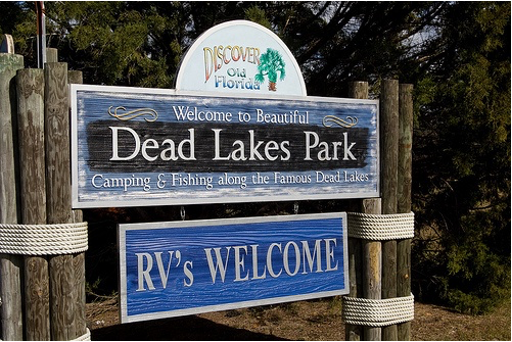}
       \caption{}
       \label{fig:intro_img}
   \end{subfigure}
   \hfill
   \begin{subfigure}{0.49\linewidth}
       \includegraphics[width=\linewidth, height=0.6\linewidth]{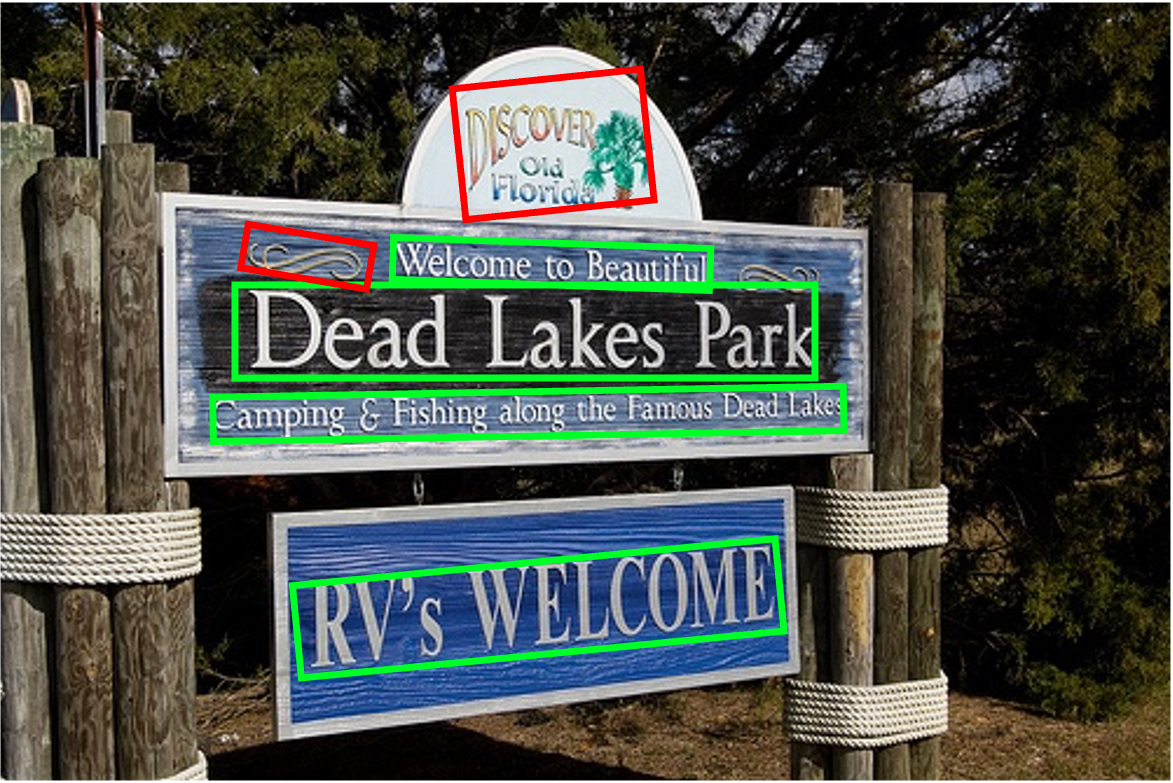}
       \caption{}
       \label{fig:intro_sem}
   \end{subfigure}
   \hfill
   \begin{subfigure}{0.49\linewidth}
       \includegraphics[width=\linewidth, height=0.6\linewidth]{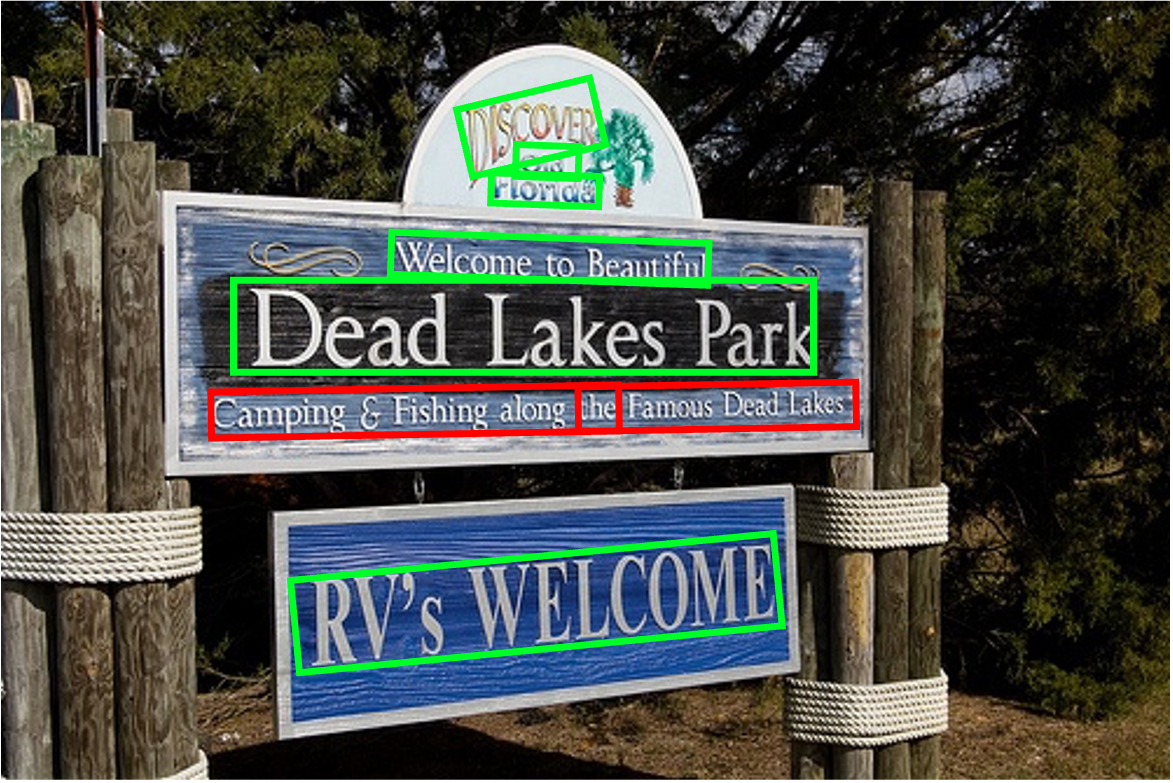}
       \caption{}
       \label{fig:intro_geo}
   \end{subfigure}
   \hfill
   \begin{subfigure}{0.49\linewidth}
       \includegraphics[width=\linewidth, height=0.6\linewidth]{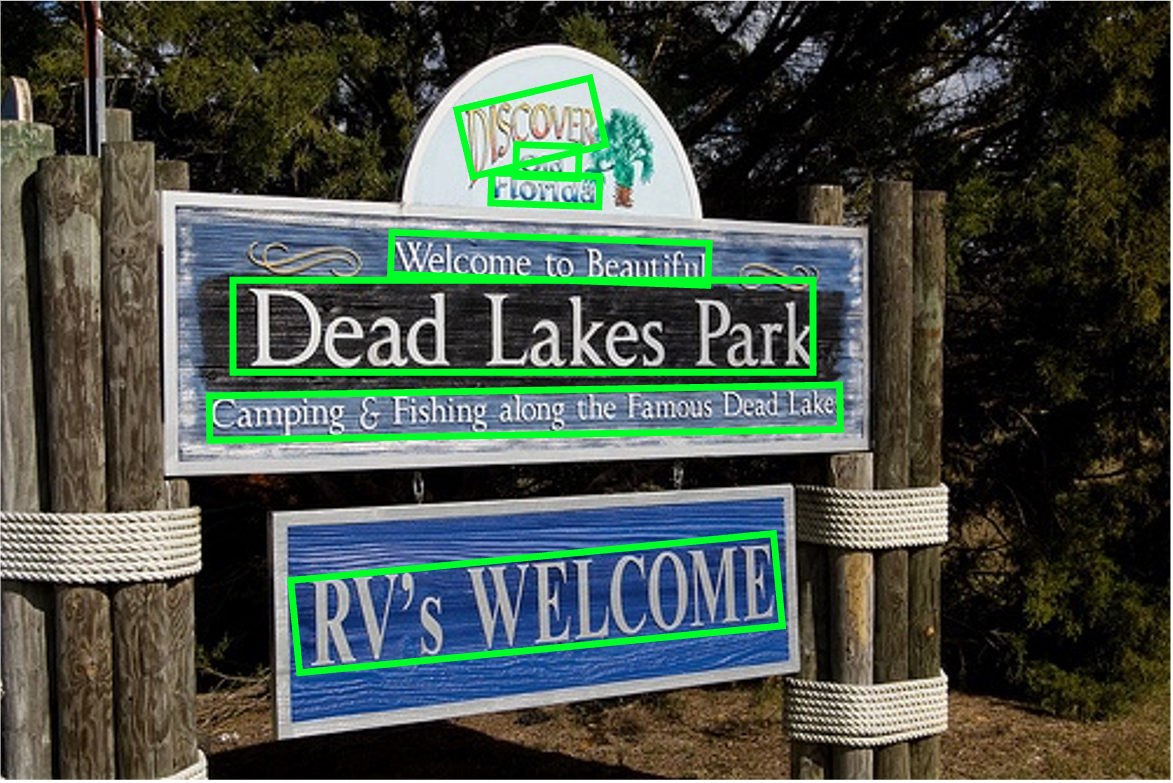}
       \caption{}
       \label{fig:intro_ours}
   \end{subfigure}
  \caption{Comparison between existing detectors and our proposed CPN. The input image is shown in (\subref{fig:intro_img}). Segmentation based methods tend to produce false positives and negatives while facing compact layouts as illustrated in (\subref{fig:intro_sem}). Anchor based approaches struggle for texts of large aspect ratio as shown in (\subref{fig:intro_geo}). Our CPN addresses the above issues by complementary proposals and features, with results given in (\subref{fig:intro_ours}). True positive results are colored in green, while the false boxes are indicated in red.}
  \label{fig:intro}
\end{figure}

Automated detection of various texts in scenes has been extensively studied for years due to its applications in many useful tasks such as multi-modal image and video understanding, autonomous indoor and outdoor navigation, etc. 
% Leveraging the recent advance of deep neural networks, quite a few scene text detection networks have been developed with very impressive performance. 
However, the detection of text in arbitrary shapes and complex layouts remains a challenge for most existing scene text detectors, leading to a high number of false positives and false negatives as illustrated in Figure \ref{fig:intro}b and \ref{fig:intro}c.

Prevalent scene text detection work follows two typical paradigms: segmentation-based approach and anchor-based approach. The segmentation approach has achieved great success thanks to its simplicity and capacity to deal with text of arbitrary shapes \cite{liao2020real, wang2019shape}. However, this approach often struggles while facing texts of complex layouts and background noise, as illustrated in Figure \ref{fig:intro}b. This is largely due to the challenges in identification of text boundaries and complicated heuristic post-processing for pixel grouping \cite{wang2019shape, deng2018pixellink,tian2019learning}. On the other hand, the anchor approach generally achieves higher accuracy due to its two-stage design with proposal generation and verification. However, it is designed for generic object detection tasks and often suffers while dealing with scene texts of arbitrary shapes such as curved texts and long text lines, as illustrated in \ref{fig:intro}c.

\begin{figure*}[t]
 \setlength{\abovecaptionskip}{0.2cm}
  \centering
  \includegraphics[width=0.9\linewidth]{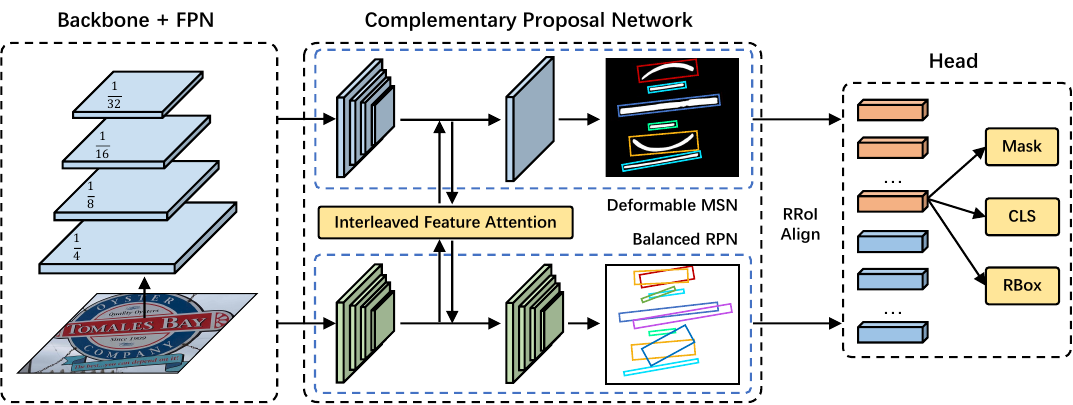}
    \caption{The pipeline of the proposed Complementary Proposal Network (CPN). Given an input image, multi-scale FPN features are extracted and fed to two parallel networks (Deformable MSN and Balanced RPN) for generating complementary semantic and geometric proposals. An Interleaved Feature Attention (IFA) module is designed to promote interaction between branch features, encouraging them to capture more spatial and scale-aware information. Features for the merged proposals are then identified with RRoI align before passing to the RoI head, where final text boxes and masks are generated.
}
\label{fig:intro_pipeline}
\end{figure*}

In light of the aforementioned analysis, we contend that the segmentation approach and the anchor approach are inherently complementary to each other, not only in terms of candidate proposals but also in the features used to generate them. By combining the strengths of these two approaches, we could potentially address their respective limitations and lead to improved performance in unconstrained scene text detection. 
However, directly integrating these two approaches faces two issues. First, the textual representation of them are distinct, how can effective and efficient unification be achieved? Second, segmentation-based methods require complex post-processing to generate candidate boxes on the CPU. This results in low efficiency for both training and inference of the entire network, preventing end-to-end training. Here, we present a \textbf{C}omplementary \textbf{P}roposal \textbf{N}etwork (CPN) that exploits rich semantic and geometric information for detecting text instances. Our CPN consists of two complementary designs as shown in Figure \ref{fig:intro_pipeline}. First, we propose two efficient complementary networks for proposal generation, namely, Deformable Morphology Semantic Network (Deformable MSN) and Balanced Region Proposal Network (Balanced RPN). Deformable MSN learns high-level semantic information with deformable morphological operators to obtain semantic proposals, while Balanced RPN captures low-level geometric structure of text instances to predict geometric proposals. Second, we introduce an Interleaved Feature Attention (IFA) module, which not only enhances the mutual information between semantic and geometric features and explicitly supervises each other, but also reduces the computational burden of the pipeline.

We conducted extensive experiments on five text detection benchmarks to evaluate the performance of our proposed CPN. Results show that the CPN approach significantly outperforms state-of-the-art methods, where it achieves an f-measure of 81.7\% (+3.6\%) on the large-scale arbitrary-shaped IC19-ArT dataset \cite{chng2019icdar2019}. Furthermore, our CPN achieves a comparable system efficiency (11.2 fps) to standard two-stage detectors such as Mask R-CNN (9.6 fps) on IC15 dataset. To summarize, the main contributions of this work are as follows:
\begin{enumerate}
    \item We propose a dual Complementary Proposal Network (CPN) that integrates semantic and geometric proposals, along with the features used to generate them. To the best of our knowledge, this is the first study to explore the integration of two distinct proposal paradigms for text detection.  
    \item We present a Deformable MSN with a deformable morphology operator to efficiently generate semantic proposals with parallelized GPU computations. 
    \item We introduce an Interleaved Feature Attention module that enables interaction and supervision between semantic and geometric features for complementary features and computation reduction. 
     \item Extensive experiments over multiple prevailing benchmarks show our proposed CPN outperforms the state-of-the-art by large margins.
\end{enumerate}

\section{Related Work}
\label{sec:related_work}

Before the deep learning era, bottom-up pipelines are widely adopted in scene text detection with hand-crafted features such as Maximally Stable Extremal Regions (MSER) \cite{neumann2012real} and Stroke Width Transform (SWT) \cite{epshtein2010detecting}. Over the past few years, deep learning based text detectors have become prevalent which can be roughly divided into anchor-based methods and segmentation-based methods.

\subsection{Anchor-based Methods}
The anchor-based methods treat text instances as common objects and adapt the pipeline of generic object detection, e.g., SSD \cite{liu2016ssd} and Faster-RCNN \cite{ren2015faster}, into text detection. They utilize offset regression from predefined anchor boxes or points in feature maps to predict text locations. To address the problem of large aspect ratios in scene texts, TextBoxes \cite{liao2017textboxes} has designed compact anchors with different aspect ratios and scales to cover texts with varied sizes. RRPN \cite{ma2018arbitrary} adjusts horizontal anchors to rotated ones with angle prediction to localize arbitrary-oriented text regions, followed by a rotated RoI pooling layer before text/non-text classification. DMPNet \cite{liu2017deep} trys to handle multi-oriented text instances with tighter quadrilateral sliding windows instead of horizontal ones. In general, anchor-based methods have two stages with proposal generation in the first stage, followed by a text verification network. Therefore, they tend to achieve better performance when compared with the simplified one-stage segmentation-based approaches.

\subsection{Segmentation-based Methods}
Methods belonging to this category draw inspiration from semantic segmentation and they aim to gather individual pixels for accurate text detection. PixelLink \cite{deng2018pixellink} predicts the linkage relationships between pixels to group pixels within the same text instance. PSENet \cite{ wang2019shape} adopts a progressive scale algorithm to expand the text kernels of different scales. In TextField \cite{xu2019textfield}, the offset field of each pixel in text regions is predicted for connecting neighborhood pixels. To effectively simplify the post-processing step, Liao \textit{et al.} \cite{liao2020real} propose differentiable binarization and achieve a good balance between speed and accuracy. To conclude, segmentation-based methods predict text boxes in one shot, making the whole pipeline much simpler and more efficient. Besides, it can intrinsically handle text of arbitrary shapes. %However, the performances of these methods are strongly affected by the quality of segmentation accuracy.

%Apart from the above two main approaches, there are also methods that try to detect parts of text regions and then link them together. SegLink \cite{shi2017detecting} is proposed to address the limitation of long words by splitting words into small segments and links. Similarly in CRAFT \cite{baek2019character}, a weakly supervised framework is proposed to predict individual character boxes and their affinity relation. %In \cite{tang2022few}, highly relevant text features are sampled and a transformer is adopted to group the features into text instances. %The intuitive advantage is that they can handle text regions with arbitrary length and orientation. 

Despite their respective superiority, methods in each group may suffer from different weaknesses. Anchor-based methods may fail when handling text instances of irregular shapes and extreme aspect ratios due to their structural limitations. Segmentation-based methods behave poorly for adjacent and overlapping text instances. %For part-based methods, either the pre-processing step to generate text parts or the post-processing step to group parts into text boxes is tedious. 
However, those approaches intrinsically compensate each other considering the above merits and demerits. Therefore, we consider proposing a novel pipeline that can combine their merits to overcome their demerits.

\section{Methodology}
\label{sec:methodology}
%-------------------------------------------------------------------------
The architecture of the proposed text detection pipeline is presented in Figure \ref{fig:intro_pipeline}. 
It consists of three components: the backbone network, the proposed Complementary Proposal Network ({CPN}), followed by an lightweight RoI head. The CPN is composed of two novel parallel complementary branches designed to efficiently integrate semantic and geometric proposals. Additionally, the Interleaved Feature Attention (IFA) module is integrated into the CPN to enhance the interactions between semantic and geometric features, further improving their complementarity.

The main objective of the CPN is to generate diverse text proposals that can accommodate a wide range of scales, shapes, and orientations under varied scenes. To this end, we have designed two networks that work in tandem to strengthen their merits and overcome their respective demerits. Specifically, the Deformable Morphology Semantic Network (Deformable MSN) generates semantic proposals, while the Balanced Region Proposal Network (Balanced RPN) generates geometric proposals. These proposals are merged as shown in Figure \ref{fig:intro_pipeline} and passed to the RROI align layer. Considering that the Deformable MSN network has already captured rich semantic information, we adopt a lightweight mask head in R-CNN by decreasing the four 3x3 convolution layers to one.
We optimize the entire pipeline through end-to-end multi-task learning without complicated post-processing steps. 

\subsection{Deformable Morphology Semantic Network}
\begin{figure}[t]
  \centering
  \includegraphics[width=1\linewidth]
    {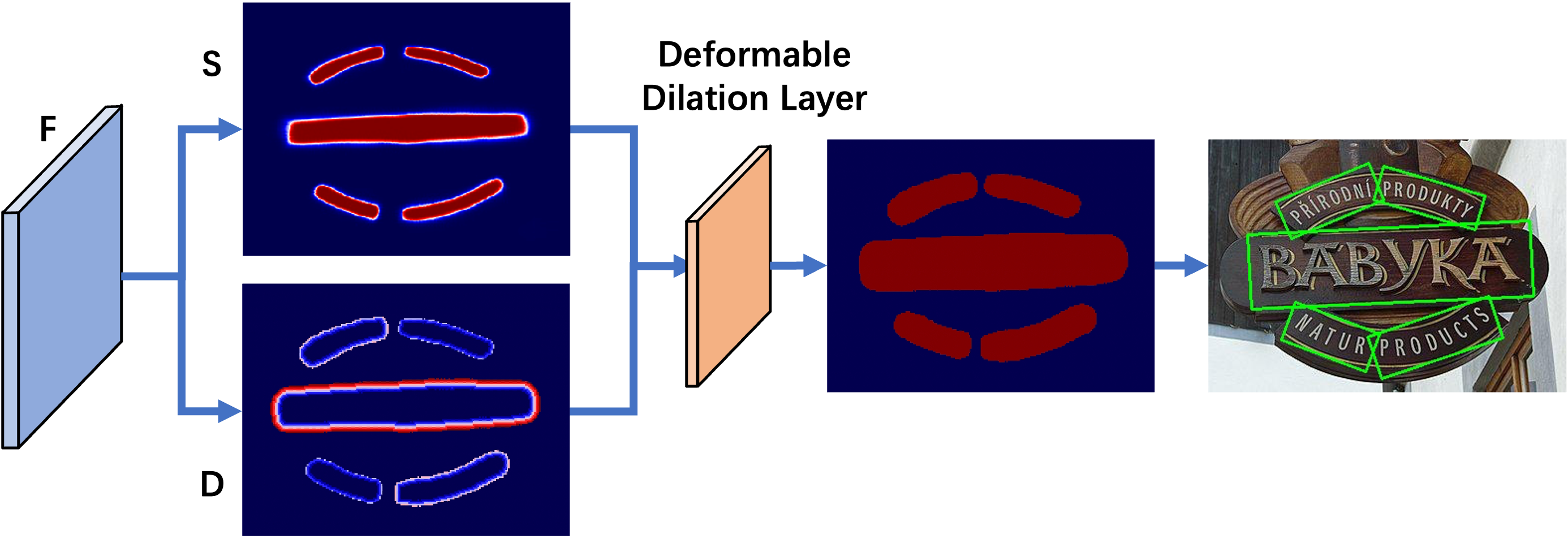}
\caption{Illustration of the proposed Deformable Morphology Semantic Network (Deformable MSN) for semantic proposals. It predicts a text erosion map $S$ and a structuring kernel map $D$, followed by a deformable dilation layer to produce text candidate regions, and thereafter the corresponding oriented proposals.}
  % \caption{Semantic proposal generation pipeline. The semantic feature predicts 2-channels maps that consist of text eroded map $S$ and structuring element size map $E$. }
  \label{fig:semantic_proposal_gen}
\end{figure}

The Deformable Morphology Semantic Network (Deformable MSN) in the upper part of the CPN in Figure \ref{fig:intro_pipeline} generates accurate candidate proposals, particularly for text with curved shapes and extreme aspect ratios where the geometric proposal branch often fails. Unlike existing segmentation-based approaches \cite{liao2020real,wang2019shape,wang2019efficient} that utilize CPU extensive operations such as the Vatti clipping algorithm \cite{vatti1992generic} for box generation, we propose a Deformable MSN with differentiable morphology operators that can be fully parallelized on GPUs, resulting in superior efficiency. Furthermore, as the Deformable MSN utilizes instance segmentation as supervision, the generated proposals can capture high-level semantic information effectively. The detailed structure of Deformable MSN is presented in Figure \ref{fig:semantic_proposal_gen}. It predicts a text erosion map and a structuring kernel map, followed by a deformable dilation layer to produce the text confidence map, and then the corresponding oriented rectangular proposals.

\noindent
{\bf Text Erosion Map Generation} In order to create labels for the text erosion map during training, we erode the text regions by specific sizes and set pixels inside as positive while others as negative. In contrast to previous approaches \cite{liao2020real,wang2019shape,wang2019efficient} which utilize Vatti clipping, we design a more efficient approach by using morphological erosion operators with circular structuring kernels of varying sizes to shrink the pixels in original text regions. For a text instance $t$, the circular structuring kernel size is given by
\begin{align}
e_t & = k \times A_t / {L_t} 
\label{eq:box_shrink}
\end{align}
where $A_t$ and $L_t$ are the area and perimeter of $t$, and $k$ is the scale hyperparameter.

\noindent
% {\bf Structuring Element Size Map Generation} For morphological operators, the structuring element is used to provide the specified size and shape information when interacting with a given image. In this work, the structuring element size map ($\mathcal{D}$) is an observation scale map, and the observation scale of the text pixel p to nearest pixel Np in the text eroded area is defined as
{\bf Structuring Kernel Map Generation} The morphological operator adopts structuring kernel to provide specified size and shape information while transforming an image. We define the structuring kernel ${D}(p)$ for a given pixel $p$ as %In this work, the structuring kernel for a given pixel $p$ is denoted as ${D}(p)$ and defined by:
%In this study, the structuring kernel serves as an observation scale map and we define the structuring kernel ${D}(p)$ for a given pixel $p$ as
\begin{align}
{D}(p)=\left\{
\begin{array}{cc}
{k} \times \left|\overrightarrow{p S_{p}}\right|, & p \in \mathbb{T} \\\relax
{0}, & p \notin \mathbb{T}
\end{array}\right.
\end{align}
where $p$ and $S_p$ belong to the same text instance and $S_p$ is the nearest pixel for $p$ in the erosion map $S$. $k$ is the normalization factor and $\mathbb{T}$ represents all the text pixels. For pixels in non-text areas ($p \notin \mathbb{T}$), we set the values to 0. 
% \begin{align}
% \mathcal{D}_{g t}(p)=\left\{\begin{array}{cc}
% \left|\overrightarrow{p B_{p}}\right| / L, & p \in \mathbb{T} \\
% 0, & p \notin \mathbb{T}
% \end{array}\right.
% \end{align}
% For the non-text area ($p \notin \mathbb{T}$), we set the values of those pixels to 0. $L$ is the normalized factor.

\noindent
% {\bf Semantic Proposal Generation} In most segmentation-based pipelines, post-processing is carefully designed over text confidence map to rebuild text instances. However, they inevitably introduce some complicated operators, e.g. Vatti-clipping algorithm, which are difficult to be paralleled on GPU and often lead to inferior performance. To build a more efficient and end-to-end trainable network, we propose the deformable morphological dilation layer which is defined as:
{\bf Semantic Proposal Generation} In most segmentation-based pipelines, post-processing is sophistically designed over text confidence map to rebuild text instances. However, they inevitably introduce some complicated operators, e.g. Vatti-clipping algorithm, which are difficult to be paralleled on GPU and often lead to inferior performance. To build a more efficient and end-to-end trainable network, we propose the deformable morphological dilation layer which is defined as
\begin{align}
\begin{aligned}
&\operatorname{dst}(x, y)  =  \max _{\left(\Delta x, \Delta y\right)} \operatorname{S}\left(x+\Delta x, y+\Delta y\right) 
% &\operatorname{dst}(x, y)  =  \max _{\left(\Delta x, \Delta y\right): \left(\Delta x, \Delta y\right) <  {\left \lceil d_{x,y} \right \rceil}^2} \operatorname{S}\left(x+\Delta x, y+\Delta y\right) 
% {\Delta x }^2+{\Delta y }^2  <  {\left \lceil d_{x,y} \right \rceil}^2
\end{aligned}
\end{align}
where $(\Delta x )^2+(\Delta y)^2  <  {\left \lceil d_{x,y} \right \rceil}^2$. $d_{x,y}$ is the predicted structuring kernel size at pixel $p(x,y)$ and $\left \lceil d_{x,y} \right \rceil$ is the corresponding radius.

Given the predicted text erosion map $S$, we binarize it and conduct connected-component labeling to obtain text instances map. Then, the labeled map is fed into the deformable morphological dilation layer with structuring kernel map $D$ to rebuild binary text instances. The deformable morphological dilation layer can be seen as a MaxPooling operator whose kernels are in circular shapes with varying radius at different locations. Finally, we produce oriented rectangular boxes by computing the rotated minimal area rectangles from the previous dilated binary regions. In general, the number of proposals generated by the Deformable MSN is less than 100, which is much smaller than the geometric proposal network. 

\subsection{Balanced Region Proposal Network} 

To address multi-oriented problem for anchor-based frameworks, existing works like \cite{ma2018arbitrary} have designed rotated anchors with different angles, scales and aspect ratios, with extensive computation. 
To efficiently generate oriented text proposals, we explore the midpoint offset representation by oriented Region Proposal Network (RPN) \cite{xie2021oriented} to capture the geometric characteristics, which complement the deformable morphology network that deals from the semantic perspective. 

\noindent
{\bf Proposal Number Balance} In most anchor-based detection pipelines, the maximum number of proposals is set to more than 1k for better performance, e.g. 2k proposals in \cite{xie2021oriented} and 1k proposals in \cite{he2017mask}. Considering that the semantic branch normally provides much less ($\sim$100) and tighter proposals covering the majority of text regions, we doubt the redundancy of the geometric proposals by RPN. Moreover, a larger number of proposals from RPN may introduce more false positives and dominate the training loss. In practice, we experiment with different numbers of proposals and the outcome is consistent with our assumption. Hence we balance the number of RPN proposals to 300 and refer to this proposal branch as Balanced Region Proposal branch (Balanced RPN). Detailed settings are described in Section \ref{subsec:ablation} with corresponding results in Table \ref{tab:proposal_influence}.

Meanwhile, we adjusted the anchor ratios to \(\{0.5, 1, 3\}\) and set the base scale to 5, making it more suitable for the scene text detection task.

\begin{figure}[t]
  \centering
  \includegraphics[width=1\linewidth]{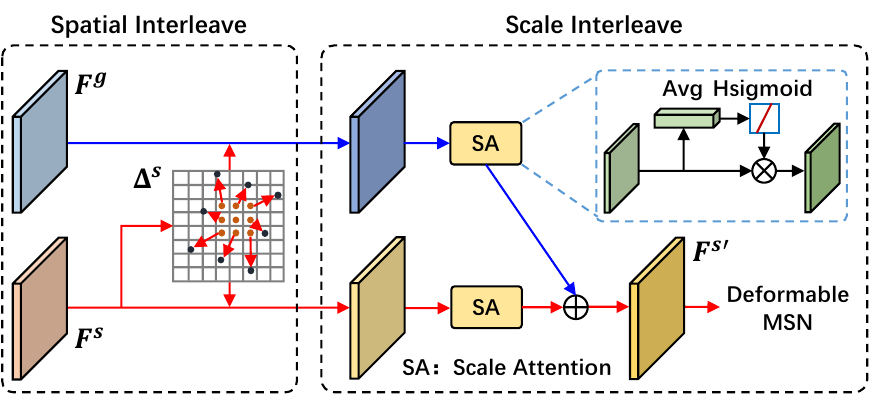}
   \caption{Detailed structure of IFA module with spatial and scale interleaved attentions for the Deformable MSN branch, and the Balanced RPN branch shares the same design.}
   \label{fig:ifa}
\end{figure}

\subsection{Interleaved Feature Attention}
\label{subsec:IFA}
To further leverage the complementary information, we propose an Interleaved Feature Attention (IFA) module between different branches for more interactions, as illustrated in Figure \ref{fig:ifa}. By interacting with each other, the semantic branch and geometric branch can encode both spatial and scaled context. Take the semantic proposal branch on the top part of Figure \ref{fig:ifa} as an example. We first apply spatial aware attention $\mathcal{O}$ among different levels of FPN features $F^s$ to obtain the offset field $\Delta^s$ given by Equation \ref{eq:ifa}. Then $\Delta^s$ will be shared to different deformable convolutions \cite{dai2017deformable} $\mathcal{D}_s$ and $\mathcal{D}_g$ to obtain the interleaved features of the semantic branch and geometric branch respectively. As the semantic features $F^s$ and geometric features $F^g$ may be in different feature spaces, we further utilize scale-aware attentions $\mathscr{S}_s$ and $\mathscr{S}_g$ to learn the corresponding weights for both features before fusion. The ultimate features $F^{s\prime}$ for the semantic branch is produced by adding the previous two features after scale attention. 

\begin{align}
    \begin{aligned}
    \Delta^s &= \mathcal{O} \left( F^s \right) \\
    F^{s\prime} &= \frac{1}{2} \sum_{i \in \{s, g\}} \left( \mathscr{S}_i \left( \mathcal{D}_i\left( F^i, \Delta^s \right) \right)\right)\\
    \end{aligned}
    \label{eq:ifa}
\end{align}

The scale aware attention $\mathscr{S}$ is given as  %in Equation \ref{eq:scale-attention}. 

\begin{align}
    \mathscr{S}\left( F \right) = \rho \left( f \left( \frac{1}{HW}\sum_{HW}F \right) \right) \cdot F
    \label{eq:scale-attention}
\end{align}

where \(H\) and \(W\) indicate the spatial scale of corresponding feature map \(F\), \(f(\cdot)\) is a linear function approximated by a \(1\times 1\) convolutional layer, and \(\rho(x)=\max\left( 0,\min\left(1,\left(x+1 \right)/2 \right) \right)\) is a hard-sigmoid function.

The proposed IFA is a lightweight module that captures geometric and semantic information to produce complementary text features, leading to an obvious performance lift in both recall and precision. Additionally, IFA deducts the original FPN features from 256 channels to 128 channels, resulting in faster inference speeds and reduced overall model complexity.

\subsection{Training Targets}
The whole pipeline is jointly trained with the loss given by
\begin{align}
\mathcal L = \mathcal L_{t}+\alpha_1 \mathcal L_{e}+\alpha_2 \mathcal L_{geo}+\alpha_3 \mathcal L_{rcnn}
\end{align}
Where \( \mathcal L_{t}\), \( \mathcal L_{e}\), \( \mathcal L_{geo}\), \( \mathcal L_{rcnn}\) represents the losses for the text erosion map, structuring kernel map, Balanced RPN and Mast R-CNN head, respectively. The $\alpha_1$, $\alpha_2$, $\alpha_3$ are the corresponding weights for each loss. 

\noindent
{\bf Loss for Text Erosion Map} To avoid the network bias to non-text pixels, we adopt dice coefficient loss for shrunk text instances in Deformable MSN. The dice coefficient $\mathcal L_{t}$ is computed as
\begin{align}
\mathcal L_t & = \frac{2 \times \sum_p\left ( \hat{S}_p \times S_p^* \right )  }{\sum_p {\hat{S}_p}^{2}+\sum_p{S_p^*}^2}  
\label{eq:seg_loss}
\end{align}
where $\hat{S}_p$ and $S_p^*$ refer to the value of pixel $p$ in the predicted and ground truth text erosion map, respectively. \\
% where $\hat{S}_p$ and $S_p^*$ refer to the value of pixel $p$ in the predicted text erosion map $\hat{S}$ and ground truth text erosion map $S^*$, respectively.
\noindent
{\bf Loss for Structuring Kernel Map} We extend the smoothed $L1$ loss proposed in \cite{girshick2015fast} by adding an extra ratio term. Then the loss function can be defined as
\begin{align}
    \mathcal L_{e} = \frac{1}{\Omega } \sum_{p\in \Omega } SL1(\hat{D}_p, D_p^*) + \\
    \beta \log \frac{\max (\hat{D}_p, D_p^*)}{\min (\hat{D}_p, D_p^*)}
\end{align}

where $\hat{D}_p$ and $D_p^*$ are the predicted and ground truth values for pixel $p$ in the structuring kernel map. $SL1()$ is the smoothed $L1$ loss and  $\beta$ is the weighted factor. $\Omega$ denotes the set of positive elements in $D_p^*$.
% where $\hat{D}_p$ is the predicted structuring kernel map $\hat{D}$ and $D_p^*$ is the value in ground truth $D^*$ at text pixel $p$. $\beta$ is the balancing weight for ratio term.
\\
\noindent
{\bf Loss for Balanced RPN} We use the oriented RPN loss \cite{xie2021oriented} to optimize the Balanced RPN branch given by
\begin{align}
\mathcal L_{geo} = \frac{1}{N} \left ( \sum_{i = 1}^{N} F_{cls}\left ( y_{i}^{*}, y_{i} \right ) + y_{i} \sum_{i = 1}^{N} F_{reg}\left ( t_{i}^{*},t_i  \right ) \right ) 
\label{eq:geo_loss}
\end{align}
% \begin{align}
% \mathcal L_{geo} = \frac{1}{N} \left ( \sum_{i = 1}^{N} F_{cls}\left ( p_i,p_{i}^{*}  \right ) + p_{i}^{*} \times \sum_{i = 1}^{N} F_{reg}\left ( t_i,t_{i}^{*}  \right ) \right ) 
% \label{eq:geo_loss}
% \end{align}
where $N$ is the number of anchors in a mini-batch and $i$ is the index. $y_{i}$ is the label and $y_{i}^{*}$ denotes the probability. $t_i$ is ground-truth offset while $t_i^*$ is the predicted one. \(F_{cls}\) is the cross-entropy loss and \(F_{reg}\) is the Smooth $L1$ loss.

\section{Experiments}
% As universal scene text includes diverse shapes, such as multi-oriented, multi-lingual, varying lengths, curved, tilted, overlapping layouts and so on, 
We adopts five widely studied datasets IC19-ArT \cite{chng2019icdar2019}, CTW1500 \cite{yuliang2017detecting}, IC17-MLT \cite{nayef2017icdar2017}, IC15 \cite{karatzas2015icdar}, MSRA-TD500 \cite{yao2014unified}, which contain a variety of different scenarios, to evaluate the performance of our proposed complementary network.

\begin{table*}[t] \normalsize
\renewcommand*{\arraystretch}{1.1}
\centering
\setlength{\tabcolsep}{8pt}
\resizebox{1.0\hsize}{!}{
\begin{tabular}{l|p{0.7cm}|p{0.4cm}p{0.4cm}p{0.4cm}p{0.5cm}|p{0.4cm}p{0.4cm}p{0.4cm}p{0.5cm}|p{0.4cm}p{0.4cm}p{0.4cm}p{0.5cm}|p{0.4cm}p{0.4cm}p{0.4cm}p{0.5cm}}
\Xhline{1.2pt}

\multirow{2}{*}{\textbf{Method}} &  \multirow{2}{*}{\textbf{Ext}} & \multicolumn{4}{c|}{\textbf{IC15}} & \multicolumn{4}{c|}{\textbf{MSRA-TD500}} & \multicolumn{4}{c|}{\textbf{CTW1500}} & \multicolumn{3}{c}{\textbf{IC19-ArT}} \\
\cline{3-17}  & & \textbf{R}   & \textbf{P}   & \textbf{F}  & \textbf{FPS} & \textbf{R}  & \textbf{P}  & \textbf{F} & \textbf{FPS} & \textbf{R}  & \textbf{P}  & \textbf{F} & \textbf{FPS} & \textbf{R}   & \textbf{P} & \textbf{F} \\
\hline
\textbf{\textit{Sem-based}} \\
\hline
PSENet-1s \cite{wang2019shape}     & MLT  & 84.5   & 86.9   & 85.7 & 1.6 & -  & -  & - & - & 79.7   & 84.8   & 82.2 & 3.9 & 52.2 & 75.9  & 61.9 \\
CRAFT \cite{baek2019character}      & MLT  & 84.3   & 89.8   & 86.9 & -  & 78.2  & 88.2  & 82.9 & 8.6 & 81.1   & 86.0   & 83.5 & - & 68.9 & 77.3 & 72.9  \\
LOMO \cite{zhang2019look}      & MLT$^+$    & 83.5   & \underline{91.3}   & 87.2 & 3.4 & -  & -  & -  & - & 76.5 & 85.7   & 80.8 & 4.4 & - & - & - \\
PCR \cite{dai2021progressive}    & MLT  & -   & -   & - & -  & 82.0  & 88.5  & 85.2 & - & 82.3   & 87.2   & 84.7 & 11.8 & 66.1 & \textbf{84.0} & 74.0 \\
DB++ \cite{liao2022real}       & Syn  & 83.9   & 90.9   & 87.3 & 10.0 & 83.3  & \underline{91.5}    & 87.2 & \textbf{29.0} & 82.8   & 87.9   & 85.3 & \underline{26.0} & -  & - & - \\
TextBPN \cite{zhang2022arbitrary}       & MLT  & -   & -   & -  & - & 84.5  & 86.6  & 85.6 & 10.7 & 83.6   & 86.5   & 85.0  & \underline{12.2} & -  & -  & - \\
TextPMs \cite{zhang2022arbitrary}     & MLT  & 84.9   & 89.9   & 87.4  & - & \underline{86.9}  & 91.0  & \underline{88.9} & 10.6 & 83.8   & 87.8   & \underline{85.8} & 9.1 & -  & -  & - \\
\hline
\textbf{\textit{Geo-based}} \\
\hline
DRRG \cite{zhang2020deep}      & MLT  & 84.7   & 88.5   & 86.6 & -  & 82.3  & 88.1  & 85.1 & - & 83.0   & 85.9   & 84.5 & - & -  & -  & - \\
Contour \cite{wang2020contournet}      & -  & 86.1   & 87.6   & 86.9 & 3.5  & -  & -  & - & - & 84.1   & 83.7   & 83.9 & 4.5 & 62.1  & 73.2  & 67.2 \\
I3CL \cite{du2022i3cl}     & Syn  & - & -  & - & -  & -  & -  & - & - & 84.5 & 87.4 & \underline{85.9} & 7.6 & 71.3 & 82.7 & 76.6 \\
FSG \cite{tang2022few}   & Syn$^+$    & \underline{87.3}   & 90.9   & \underline{89.1} & \textbf{12.9}  & 84.7  & 91.4  & 87.9 & - & 82.4   & \underline{88.1} & 85.2 & - & -  & - & - \\
DPText-DETR \cite{ye2023dptext}   & MLT$^+$  & -   & -   & - & -  & -  & -  & - & - & \underline{86.2}   & \textbf{91.7} & \textbf{88.8} & - & \underline{73.7}  & 
 83.0 & \underline{78.1} \\
\hline
\textbf{\textit{Ours}}   & MLT  & \textbf{89.2} & \textbf{91.7} & \textbf{90.4} & \underline{11.2} & \textbf{88.3}  & \textbf{91.6}  &  \textbf{89.9}  & \underline{13.3}  & \textbf{89.6} & 88.0   & \textbf{88.8} & 12.0 & \textbf{79.9} & \underline{83.6}   & \textbf{81.7} \\
\Xhline{1.2pt}
\end{tabular}
}
\caption{Experimental results on IC15, MSRA-TD500, CTW1500 and IC19-ArT. ``Sem-based" represents segmentation-based methods and ``geo-based" refers to the anchor-based approaches. ``Ext" means extra data is used for pre-training. ``Syn" and ``MLT" denote the SynthText and IC17-MLT datasets, ``$^+$" denotes the use of extra data. R, P, F and FPS refer to recall, precision, f-measure and frame per second, respectively. Best results are in bold, while the second ones are underlined.}
\label{tab:res_multiori_curve}
\end{table*}

\subsection{Implementation Details}
We use ResNet50 \cite{he2016deep} as our backbone. All the networks are optimized with AdamW \cite{loshchilov2017decoupled} with batch size set to 16. On the IC17-MLT dataset, we train the model for 75 epochs without using extra data such as SynthText. The initial learning rate is set to \(1\times10^{-4} \) and divided by 10 at 65 and 70 epochs. For the rest of the datasets, we fine-tune the model with their corresponding train sets on the previous IC17-MLT model. During fine-tuning, the model is trained for 24 epochs with an initial learning rate set to \(5\times10^{-5} \) and decayed by 0.1 after 20 epochs.
Three augmentation schemes are implemented for training: 1) each side of the images is randomly re-scaled within the range of [480, 2560] without maintaining the aspect ratio, 2) each image is randomly flipped horizontally and rotated within the range of \(\left [ - 10^{\circ}, 10^{\circ} \right ]\), 3) \(640\times 640\) random samples are cropped from each transformed image. For straight-text datasets, we directly used the predicted oriented bounding boxes as detection outputs. For curved-text datasets, we use the predicted masks as detection outputs.

\subsection{Comparison with the state-of-the-arts}

As presented in Table \ref{tab:res_multiori_curve} and \ref{tab:res_multilingual}, our proposed approach demonstrates superior performance compared to previous state-of-the-art methods, with a significant margin across all five datasets and in various scenarios.

\noindent
\textbf{Curved text detection}
The dataset IC19-ArT has a balanced distribution of all text shapes, providing a comprehensive evaluation that is presented in Table \ref{tab:res_multiori_curve}. Without whistles and bells, our network outperforms the state-of-the-art by a significant margin (\textbf{3.6\%}) and achieves an impressive f-measure of 81.7\%. It's noteworthy that our approach achieves a considerable recall gain of \textbf{6.2\%} over the previous SOTA, which adopts a transformer-based pipeline. 
The experiments demonstrate that our complementary network can effectively boost both recall and precision.

\begin{figure*}[t]
 \setlength{\abovecaptionskip}{0.2cm}
  \centering
  \includegraphics[width=0.24\linewidth, height=0.17\linewidth]{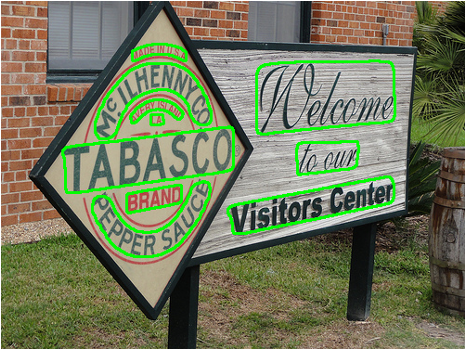}
  \includegraphics[width=0.24\linewidth, height=0.17\linewidth]{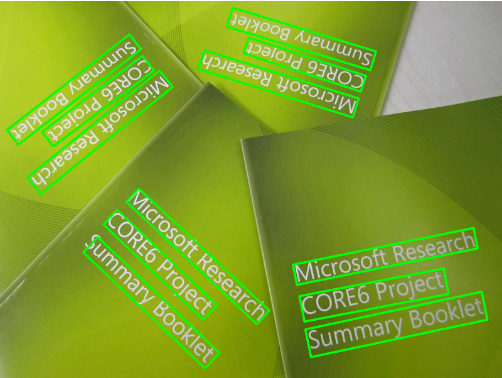}
  \includegraphics[width=0.24\linewidth, height=0.17\linewidth]{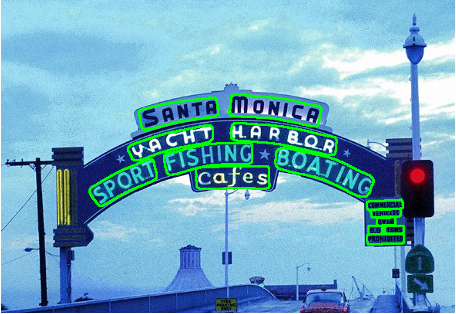}
  \includegraphics[width=0.24\linewidth, height=0.17\linewidth]{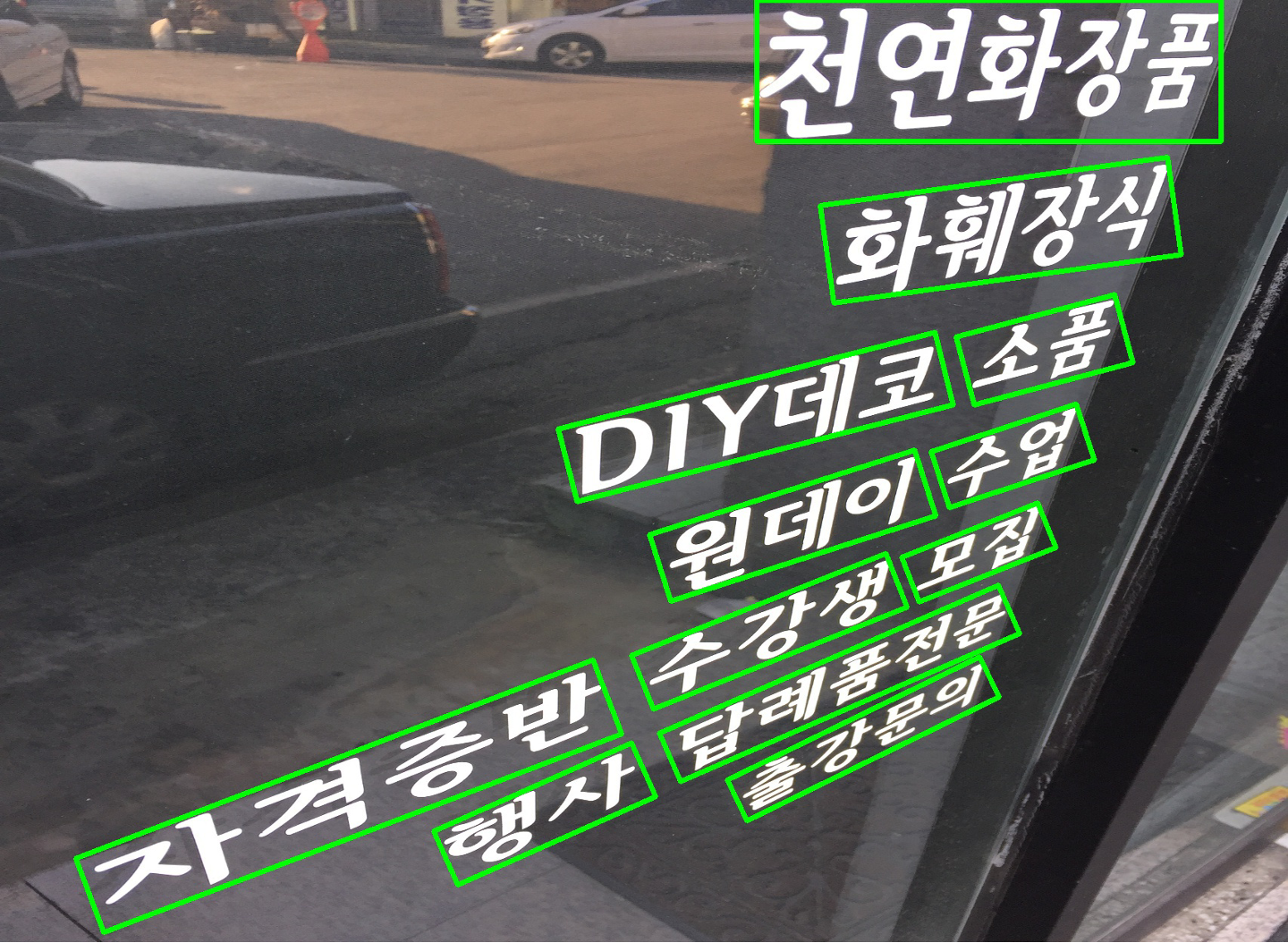} \\
  \includegraphics[width=0.24\linewidth, height=0.17\linewidth]{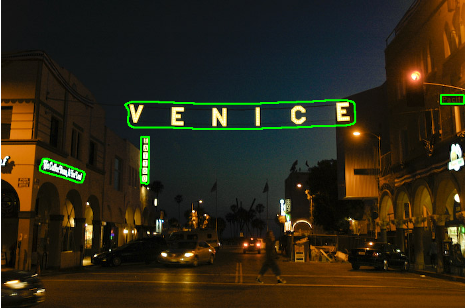}
  \includegraphics[width=0.24\linewidth, height=0.17\linewidth]{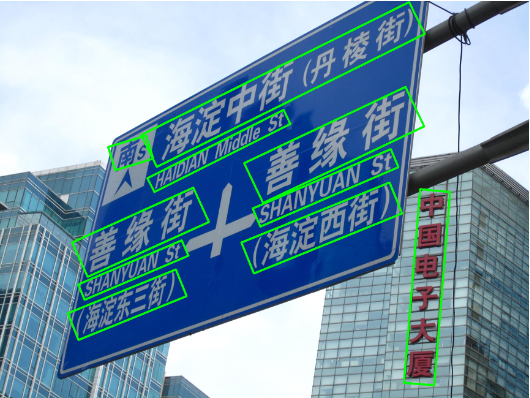}
  \includegraphics[width=0.24\linewidth, height=0.17\linewidth]{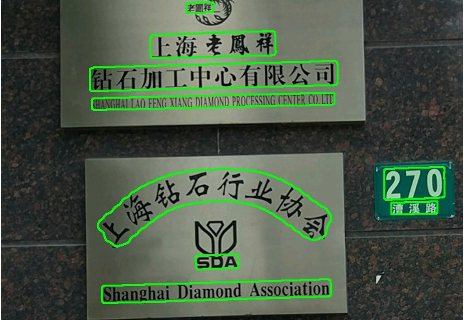}
  \includegraphics[width=0.24\linewidth, height=0.17\linewidth]{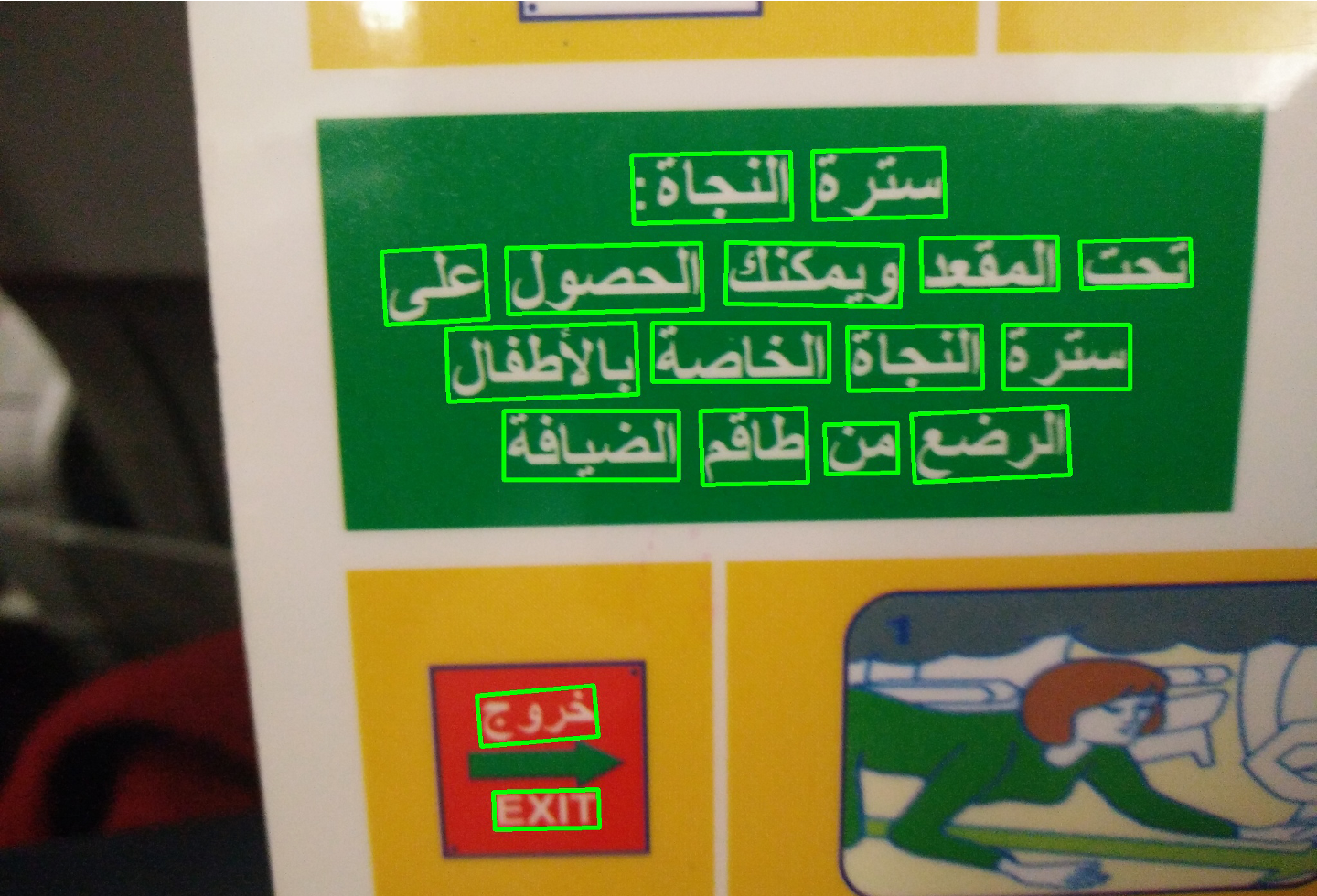}
  \caption{The qualitative results of our proposed method under various scenes, such as curved text, long text line, multi-oriented text, complex layout and multilingual text. }
%   Prediction examples of our model, on CTW1500, MSRA-TD500, Total-Text, IC17-MLT and IC19-ArT samples.
\label{fig:quality_results}
\end{figure*}

\noindent
\textbf{Multi-oriented text detection}
To validate our network on multi-oriented text instances, we evaluate on the TD500 and IC15 datasets, and the results are presented in Table \ref{tab:res_multiori_curve}. 
During inference, we resize the shorter side of the images to 960 on the IC15 dataset. 
Despite the presence of a large number of small and low-resolution text regions in IC15, our approach achieves a promising 90.4\% f-measure which surpasses state-of-the-art. For example, it outperforms FSG \cite{tang2022few} by 1.3\% on f-measure. On the TD500 dataset, we resize the longer side of images to 800 while keeping the aspect ratio. Our approach exceeds the best-reported results from TextPMs \cite{zhang2022arbitrary} by 1.0\% for f-measure with better fps. As visualized in the second column of Figure \ref{fig:quality_results}, our CPN can handle multi-oriented text instances as well as long text lines with ease.

\noindent
\textbf{Multilingual text detection}
We demonstrate the capability of our proposed approach in detecting multilingual texts on the IC17-MLT dataset. During inference, we keep the aspect ratio and resize the longer side of the images to 1920. As listed in Table \ref{tab:res_multilingual}, our network achieves an impressive f-measure of 80.0\% with a recall of 75.4\% on this challenging dataset with multilingual text. Compared to the previous state-of-the-art method, FSG \cite{tang2022few}, we achieve a 0.4\% gain in f-measure and 2.2\% in recall without using additional training data such as SynthText \cite{gupta2016synthetic}. To our best knowledge, our approach is the first framework to achieve an f-measure over 80\%.

% results on IC17-MLT dataset
\begin{table}[]
\renewcommand*{\arraystretch}{1.05}
\centering
\setlength{\tabcolsep}{9pt}
\resizebox{1\hsize}{!}{
    \begin{tabular}{l|ccc}
     \Xhline{1.2pt}
    \textbf{Method}       & \textbf{R}         & \textbf{P}        & \textbf{F}         \\
    \hline
    PSENet-1s \cite{wang2019shape}       & 68.2          & 73.8          & 70.9          \\
    LOMO \cite{zhang2019look}            & 60.6          & 78.8       & 68.5    \\
    CRAFT \cite{baek2019character}   & 68.2          & 80.6          & 73.9     \\
    SPCNet \cite{xie2019scene}         & 68.6          & 80.6          & 74.1    \\
    DRRG \cite{zhang2020deep}         & 61.0          & 75.0          & 67.3    \\
    SD \cite{xiao2020sequential}         & 72.8          & 84.2          & 78.1          \\
    DB++ \cite{liao2022real}          & 67.9          & 83.1          & 74.7     \\
    FSG \cite{tang2022few}          & \underline{73.2}          & \textbf{87.3} & \underline{79.6}    \\
    \hline
    \textbf{Ours}     & \textbf{75.4} & \underline{85.3}    & \textbf{80.0}   \\
     \Xhline{1.2pt}
    \end{tabular}
}
\caption{Experimental results on IC17-MLT dataset. }
\label{tab:res_multilingual}
\end{table}

\begin{table}[]
\renewcommand*{\arraystretch}{1.1}
\centering
\setlength{\tabcolsep}{10pt}
\resizebox{\columnwidth}{!}{%
\begin{tabular}{p{0.2cm}p{0.2cm}p{0.3cm}|p{0.3cm}p{0.3cm}p{0.5cm}|p{0.3cm}p{0.3cm}p{0.5cm}}
% \begin{tabular}{ccc|ccc|ccc}
\Xhline{1.2pt}
\multirow{2}{*}{Sem} & \multirow{2}{*}{Geo} & \multirow{2}{*}{IFA} & \multicolumn{3}{c|}{IC17-MLT} & \multicolumn{3}{c}{CTW1500} \\
\cline{4-9} &   &   & R    & P    & F    & R    & P    & F    \\
\hline 
\checkmark &   &   & 64.7 & 82.9 & 72.7 &    84.3  &   \textbf{90.1}   &   87.1   \\
  & \checkmark &   & 73.0 & 85.1 & 78.6 &   86.4   &   88.3   &   87.3   \\
\checkmark & \checkmark &   & 74.9 & 84.6 & 79.5 &  87.7    &   88.0  &  87.9    \\
\checkmark & \checkmark & \checkmark & \textbf{75.4} & \textbf{85.3} & \textbf{80.0} & \textbf{89.6} & 88.0 & \textbf{88.8} \\
\Xhline{1.2pt}
\end{tabular}%
}
\caption{Ablation study of different modules on IC17-MLT and CTW1500 datasets. “Sem” represents the Deformable MSN generating semantic proposals and “Geo” refers to the Balanced RPN producing geometrical ones.}
\label{tab:ablation_modules}
\end{table}

\begin{figure}[t]
 \setlength{\abovecaptionskip}{0.2cm}
  \centering
  \includegraphics[width=1\linewidth]{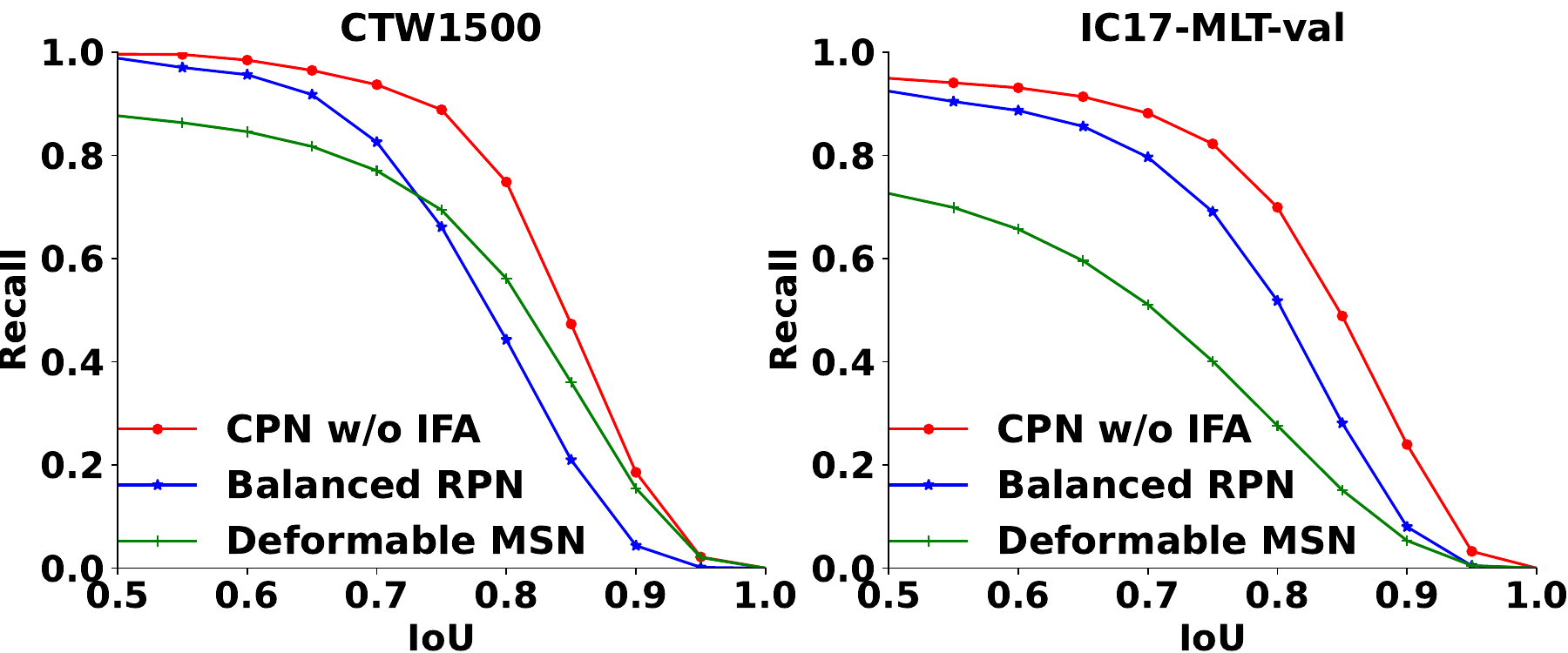}
   \caption{Recall vs. IoU overlap ratio on the CTW1500 test set and IC17-MLT validation set. Rotated bounding boxes are applied while computing IoU on CTW1500.}
   \label{fig:proposal_recall_cmp}
\end{figure}

\subsection{Complementary Proposal Analysis}
Furthermore, we investigate whether the two branches in CPN, Deformable MSN and Balanced RPN, can produce high-recall and high-quality proposals, as well as whether they complement each other. Experiments are typically conducted on the curved CTW1500 test set and the multilingual IC17-MLT validation set. In Figure \ref{fig:proposal_recall_cmp}, we give the Recall-to-IoU curve which is related to the quality of proposals. The left plots on CTW1500 test set in Figure \ref{fig:proposal_recall_cmp} show that the Deformable MSN performs better when IoU gets larger, while Balanced RPN behaves completely opposite. This is intuitively accepted since semantic proposals are more accurate than geometric proposals for curved and long text instances. Regarding to the CPN$^*$ (without the IFA module), the recall is consistently much higher than either one across all IoU thresholds, indicating that the two parallel branches can effectively complement each other. On the IC17-MLT validation set, Balanced RPN gives higher recall than Deformable MSN under all IoU threshold settings, suggesting that anchor-based methods may perform better than segmentation-based methods on challenging multilingual datasets. Similarly, our CPN$^*$ shows a steady increase in recall compared to both approaches. In summary, the proposed two networks can well complement each other with much higher recall.
% Additional experiments are given in the supplementary to show that the CPN$^*$ design can boost the performance of either proposal network via multi-task learning and IFA.

\subsection{Ablation Study}
\label{subsec:ablation}

\noindent
\textbf{Effectiveness of the two proposal networks} 
To validate the effectiveness of the proposed Deformable MSN and Balanced RPN branches, we conduct ablation study on the IC17-MLT and CTW1500 datasets. The components studied and corresponding results are summarized in Table \ref{tab:ablation_modules}. With regards to the Deformable MSN, the recall is much lower compared to the Balanced RPN, which validates the proof in Figure \ref{fig:proposal_recall_cmp}. The consistent and substantial gains observed on both datasets after integrating the two networks demonstrate the effectiveness of our design.

\noindent
\textbf{Effectiveness of the IFA}
Table \ref{tab:ablation_modules} shows that the network can further improve the f-measure by 0.5\% and 0.9\% on the IC17-MLT and CTW1500 datasets when incorporating the IFA. This suggests that the interaction between semantic and geometric features is crucial for enhancing performance. Furthermore, the IFA employs a lightweight design with only half the number of feature channels of FPN, which reduces the overall network's GLOPs by about 15\% and parameters by 12\%, as illustrated in Table \ref{tab:model_complexity}.

\begin{table}[]
\centering
\resizebox{\columnwidth}{!}{%
\begin{tabular}{l|cccccc}
\Xhline{1.2pt}
\multirow{2}{*}{\textbf{Dataset}} & \multicolumn{6}{c}{\textbf{Number of Geometric proposals}}                                                                                                                      \\
  \cline{2-7}     & \multicolumn{1}{c}{\textbf{0}}     & \multicolumn{1}{c}{\textbf{100}}   & \multicolumn{1}{c}{\textbf{300}}   & \multicolumn{1}{c}{\textbf{500}}   & \multicolumn{1}{c}{\textbf{1000}}  & \multicolumn{1}{c}{\textbf{2000}}  \\
\hline
IC15                     & \multicolumn{1}{c}{88.15} & \multicolumn{1}{c}{90.21} & \multicolumn{1}{c}{90.36} & \multicolumn{1}{c}{90.40} & \multicolumn{1}{c}{90.41} & \multicolumn{1}{c}{90.41} \\
CTW1500                  & \multicolumn{1}{c}{87.63} & \multicolumn{1}{c}{88.70} & \multicolumn{1}{c}{88.80} & \multicolumn{1}{c}{88.80} & \multicolumn{1}{c}{88.76} & \multicolumn{1}{c}{88.71} \\
\Xhline{1.2pt}
\end{tabular}%
}
\caption{F-meansure on the CTW1500 and IC15 dataset when setting various number of proposals in Balanced RPN.}
\label{tab:proposal_influence}
\end{table}

\begin{table}[]
\centering
\setlength{\tabcolsep}{9pt}
\resizebox{\columnwidth}{!}{%
\begin{tabular}{cl|ccc}
\Xhline{1.2pt}
\multicolumn{2}{c|}{\textbf{Method}}  & \textbf{GFLOPs} & \textbf{Params} & \textbf{FPS} \\
%\textbf{Method}  & \textbf{GFLOPs} & \textbf{Params} & \textbf{FPS} \\
\hline
\multicolumn{2}{c|}{Mask RCNN}              & 142.39 & 43.75M  &  9.6   \\
\hline
\multirow{2}{*}{Ours} & CPN w/o IFA                  & 117.74 & 42.54M   &  10.1   \\
                      & CPN    & \textbf{99.61} & \textbf{37.38M}  &  \textbf{11.2}  \\
\Xhline{1.2pt}
\end{tabular}%
}
\caption{Comparisons on the FLOPs, number of parameters, and inference speed.}
\label{tab:model_complexity}
\end{table}

\noindent
\textbf{Influence of the proposal number} 
Traditionally, anchor-based detectors require more than 1k proposals in RPN for higher recall. However, our semantic network generates proposals after the component grouping operation, which typically results in less than 100 proposals. As we have two complementary proposal networks, we investigated how the number of proposals influences detection performance by varying the number of geometrical proposals. Our results in Table \ref{tab:proposal_influence} suggest that using a large number of proposals only slightly improves performance. With the aid of Deformable MSN, our CPN becomes less reliant on redundant proposals.

\noindent
\textbf{Complexity of the model}
To analyze the model complexity of our network, we compute the FLOPs, number of model parameters, and inference speed. For a fair comparison, we resize the input images to 640 on both sides for all models to calculate the FLOPs. We also use the images from the IC15 test dataset to measure the inference speed by FPS. As shown in Table \ref{tab:model_complexity}, our proposed architecture has a lower computational cost in terms of FLOPs, model size and a faster inference speed than the standard Mask R-CNN. %Notably, the detection efficiency is evidently improved when equipped with the IFA module.

\section{Conclusions} 
We present the Complementary Proposal Network (CPN), an innovative approach that combines the strengths of segmentation-based and anchor-based methods for scene text detection. The CPN includes two efficient networks, the Deformable MSN and Balanced RPN, which work together to generate semantic and geometric proposals. The Deformable MSN produces semantic proposals based on instance segmentation, while the Balanced RPN generates geometric proposals based on pre-defined anchors. By working in concert, these two proposal branches reinforce each other's strengths and compensate for their individual weaknesses.
% The CPN also incorporates the Interleaved Feature Attention (IFA) module, which facilitates interaction between semantic and geometric features to further improve both precision and recall.
The CPN has been tested on five popular scene text detection datasets and has achieved impressive results. Our study aims to encourage further exploration of the complementary relationships between anchor-based and segmentation-based methods for scene text detection and object detection in general.

\bibliography{aaai24}

\end{document}